# Literature on Hand GESTURE Recognition using Graph based methods

Neha Baranwal and Varun Sharma

**Abstract:** Skeleton based recognition systems are gaining popularity and machine learning models focusing on points/joints in a skeleton have proved to be computationally effective and application in many areas like Robotics. It is easy to track points and thereby preserving spatial and temporal information, which plays an important role in abstracting the required information, classification becomes an easy task. In this paper, we aim to study these points but using a cloud mechanism, where we define a cloud as collection of points. A point cloud is a collection of data that represents objects or space. These points represent a single point on a sampled surface as its X, Y, and Z coordinates. The purpose of a point cloud is to group together many spatial measurements to represent a larger spatial picture. They take on 4D aspects when red, green, and blue information is added, an extra layer which can be added to any gesture to differentiate further. However, when we add temporal information, it may not be possible to retrieve the coordinates of a point in each frame and hence instead of focusing on a single point, we can use k-neighbors to retrieve the state of the point under discussion. Our focus is to gather such information using weight sharing but making sure that when we try to retrieve the information from neighbors, we do not carry noise with it. LSTM which has capability of long-term modelling and can carry both temporal and spatial information, we have used LSTM as our base algorithm and using point clouds as input have performed many experiments to find the best hyper-parameters and at the same time document the results. Our focus was to increase the accuracy of the model and study the effect of optimizers on how they play a role in gesture recognition classification. Our methodology combines both current and previous feature information using neighboring points with appropriate weight sharing and fed to LSTM network where these features are learned, and algorithm was tuned to achieve state-of-art. The proposed methodology was trained and tested on a challenging SHREC'17 dataset.
Point Clouds are not bound to a particular action or gesture and the tuned algorithm should be applicable on any action recognition and not just hand gesture recognition. However, as the density of point cloud increases, the noise carried from one frame to another can increase and filtering noise will be a necessity. We have studied methods on how the grouping of points should be done to make sure minimum noise is transferred. We in this study restrict ourselves to capture the results only on Hand Gesture SHREC'17 dataset but believe this proposal will benefit further experiments. Finally, the normalized results achieved on SHREC'17 were compared against other graph-based techniques to evaluate the effectiveness of point cloud-based methods.

**Introduction:** Hand gesture recognition started many decades ago, basically for computer control. Glove-based control interfaces in 1980s were the first steps in this area and with this it was realized that gestures inspired by sign languages[1] [3] [4] [2] [7] [8] can be used for issuing commands to computer interface. In the 1990s, there was significant increase in work for identifying gestures in images and video using computer vision. Over past decades many inventions fuelled discovery in gesture recognition, what started with wired technology, has now changed into wireless and with many areas where Hand Gesture[6][9][10] are widely used and are becoming part of life. Currently it is an open knowledge area and requires lot of studies and experiments to get better results not just accuracy but also computational cost as training such data requires lots of resources.

Approaches utilizing different methods specially using skeleton are increasing as they require less processing comparing with full image and hence less computational cost, another benefit of using such techniques is application in Robotics[11][12][13][14]. Earlier methods utilized full image and stand-alone methods of using Machine Learning, Deep Learning did give good results but were limited and training and last gesture classification was difficult. As technology developed, cameras which play an important role also got better, resulting in images with lots of modularity. Despite good quality image and easy access to GPU/TPU, this area remains an area where lots of study can be done, newer approaches are taking us towards better accuracy but there are noises which restrict their accuracy and hence they limit use of these technologies in sensitive areas like medical (surgery).

Gesture can be defined into five categories:
• Gesticulation: Spontaneous movements of the hands and arms that are accompanied by speech.
• Language-like gestures: Gesticulation integrated into a speech, replacing a particular spoken word or phrase.
• Pantomimes: Gestures that depict objects or actions, with or without accompanying speech.
• Emblems: Gestures like "V for victory", "thumbs up" and assorted rude gestures.
• Sign languages: Well defined Linguistic systems such as American Sign Language. Explained below are alphabets "A", "C" and "F".

Gesture recognition is the process by which gestures made by the user are made known to the system. It can also be explained as the mathematical interpretation of a human motion by a computing device. Various types of gesture recognition technologies in use currently are:

• **Contact type**
It involves touch-based gestures using a touch pad or a touch screen.

• **Non-Contact**
o Device Gesture Technologies - Device-based techniques use a glove, stylus, or other position tracker, whose movements send signals that the system uses to identify the gesture.

• Vision-based Technologies
There are three approaches to vision-based gesture recognition.
o Model based techniques - They try to create a 3D model of the user's hand and use this for recognition.
o Image based methods - Image-based techniques detect a gesture by capturing pictures of a user's motions during a gesture.
o Electrical Field Sensing - Proximity of a human body or body part can be measured by sensing electric fields.

Methods involving RGB data, have proved to achieve state-of-art but methods involving point clouds have proven further beneficial as they are able to capture latent geometric structure and the distance information between object surfaces. However, with limited techniques it was a challenge to utilize this information. Along with capture this information it is also required that we can capture long term relationships and LSTM does fit very well. The LSTM architecture is such that it allows long term modelling and hence increase the bar. With LSTM it is possible to capture both spatial and temporal information, during an action pose and time are two factors which describes what action is being taken. LSTM at the same time also allows to test different configurations and can be complex enough to solve any problem with higher accuracies. With point clouds capturing information is also a complex task. We can visualize point clouds in two ways, one where a point

'P' in frame 1 also available in frame '2', during this exact information we can extract the spatial-temporal information. Assuming that we can capture all points in one frame to another, task of classifying gesture can be very easy, however this is more of hypothetical scenario, and we need to look further. As we know that neighbouring points to some extent carries same information, utilizing this fact information can be extracted in another frame using the neighbouring points but this is going to be computationally very expensive. Moving a layer up, if we can average out the points in a frame and use that as information points can be computationally easy. Point clouds along with LSTM help in achieving this framework.

Having studied the current trending methods, we found that with each study we are able to device a method which can give better accuracy. The studies have claimed to be better over the other but is found that those studies focus on the architecture and devising a new way to achieve state-of-art. During this, we tend to forget small components which if tuned can help overall algorithm to give better results. There are very minimal studies on small components used in any algorithm which proved to be better than others. There is no documentation of effects of things like hyper-parameter tuning, what happens if we use optimizer A compared to optimizer B in a deep learning model. Also, whether a particular technique is better than other is missing.

With all gaps found in studies, we decided to focus on Point Clouds and LSTM as base algorithm and try to tune the current algorithm based on LSTM and point clouds. At the same time, try to capture the required information from point cloud and see using which methodology noise can be filtered out. When we use neighbours to extract the point information, we indirectly also extract information which is not related to the action (gesture). We will study the weighing mechanism so that we can analyse depending on dataset how to filter the noise. At last, we will compare our results with other GCN based methods, to conclude which methodology has an upper hand.

**Problem Statement**

Availability of fast processors like GPU, TPU, immense storage capacity and scalability has enabled experimentation on subject like action/gesture recognition. Data collection in different forms is yet difficult but not anymore, a challenging task. Different studies on Hand gestures have been done and many more are in progress, with each research bar is increased and it also paves a path for further studies on what can be and should be done. Our problem deals in area of Hand gesture recognition, after a rigorous review of existing studies we have found that few areas need more focus and results should be ready for any future task to be undertaken.

One of the problems is lot of studies done are focused on achieving state-of-art or do better than the existing best, this leads to different methods, but missing point here is, can the existing algorithm or method be enriched with small study like effects on accuracy due to hyper-parameter tuning. Are most of scenarios related to tuning are studied and documented. We found that not much has been explored on this and hence one of our focus points is to try training the model with different hyperparameters and document the changes we observe in the result.

Another problem, with methodology using Point Clouds is that, either they are computationally very expensive as they try to capture details of each point in one frame to another OR methods using k-neighbor points to find the start of point have problems where they carry noise with them. Though weighing mechanism is helpful but not enough has been tried and hence it is too early to say that we have reached a pinnacle of effort. We would like to study different ways of weighing the neighboring point clouds and document how these methods have performed on various datasets. Finally, many methods claiming to be better than the other but is missing any comparison.

We would like to normalize our accuracy matrix from confusion matrix and compare the results with other graph-based methods.

**Aim and Objectives**
The main aim of this research is to tune the cloud-based point LSTM algorithm to classify the hand gesture. The classification using cloud-based point LSTM allows use of previous and current point with less computational cost and the algorithm can also be used on other action recognition.
The research objectives are formulated based on the aim of this study, which are as follows:
• To analyze the current algorithm and tune the various hyperparameters to improve the accuracy of classification, document the effect on model performance due to various optimizers.
• To study different neighborhood grouping methodology which can help in avoiding irrelevant information (noise) to be carried further in network thereby making algorithm generic to be used on other action classification with higher accuracy.
• To compare the model results with other graph-based techniques to identify the better algorithm.

**Literature Survey:**

In this century, backed up by fast changing and improved information technologies and communication systems, new systems are being developed which helps provide better user experience and replicate real world scenarios. One such field which is continuously changing, and getter better is gesture recognition. New methodologies are being developed to enhance the recognition and Point Cloud is an effort towards same. With launch of cheap depth sensors like Kinect, Intel RealSense etc. lot of time and investment has been put to derive an algorithm which can help solving the gesture recognition, use of Point Cloud is a step towards it. However, before the point clouds be used , it is important to assess the quality of point clouds (Epfl, 2018).
Point clouds are datasets that represent objects or space. These points represent the X, Y, and Z geometric coordinates of a single point on an underlying sampled surface. Point clouds are a means of collating many single spatial measurements into a dataset that can then represent a whole. When colour information is present, the point cloud becomes 4D.
Point clouds are most generated using 3D laser scanners and LiDAR (light detection and ranging) technology and techniques. Here, each point represents a single laser scan measurement. These scans are then stitched together, creating a complete capture of a scene, using a process called 'registration'. Conversely, point clouds can be synthetically generated from a computer program. Point clouds have information hidden such that they can be used precisely describe the geometric structure and help in finding the distance between object surfaces, these informational elements provide necessary details which are useful in gesture recognition. Since point cloud is not a visual rather geometric representation of an object, it given immense possibility to build a concept which can be used for all types of gesture recognition, however our focus in this study is limited to hand gestures. 3D representation of an object help extracts both motion and structure feature and gives a possibility to derive methodologies which can help tune the processing. Below Figure 1 (Apostol et al., 2014), is an example of how point clouds looks with four different gestures.

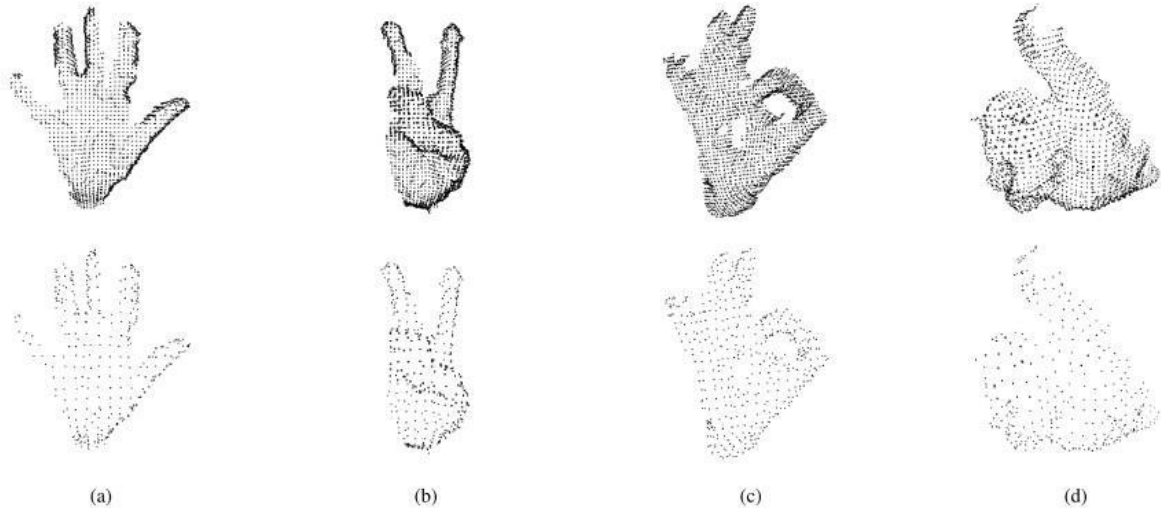
Figure 1 Gesture point clouds: (a) Open hand pose. (b) Peace pose. (c) Ok pose. (d) Like pose

**Recent Gesture Recognition techniques**
With ever evolving technology, new methods are being developed and tested in the field of Gesture Recognition. The Gestures can be divided into three sections,
• Static 2D – simplest form and shape is basically used to identify gesture example fist or fingers. They are simpler poses complexities when tracking is required[24][25].
• Dynamic 2D – It's enhancement to Static 2D recognition, where hand trajectories different features and their various combinations are used.
• 3D – Kinect sensor which allowed to capture depth in an image brough a revolution, depth map helps in identifying distance of all pixels from surface of the image, with many advantages of colour images.
This section aims to detail down the latest Gesture Recognition techniques, pros, and cons of same.

**Vision Based Gesture Recognition**
Considering the advantages of 3D images, many studies were done on 3D images and methodologies were evolved to get better accuracy.
2019 (Zhu et al., 2021), 3D shape context was used to represent 3D hand gestures. This basically was a technique to gather image information utilizing local shape context and global shape distribution of each 3D point. Once hand gestures are constructed using these 3D shape context, dynamic time warping algorithm was used to identify the gesture. Figure 2 (Zhu et al., 2021), shows pictorial representation of the method.

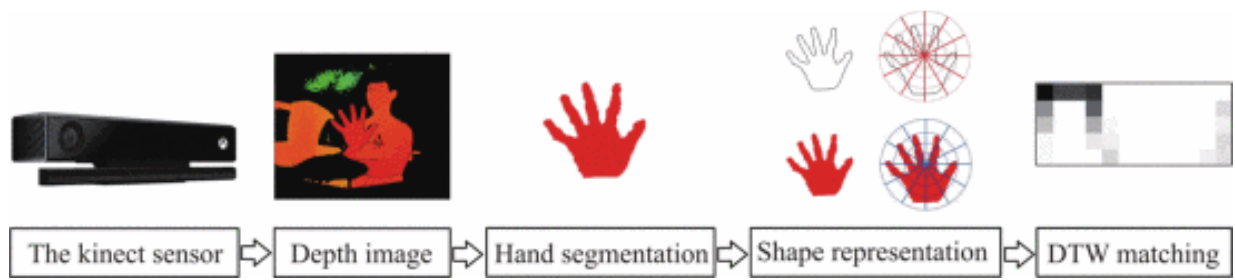
Figure 2 Representation of hand gesture using histogram and DTW algorithm

Advantages of this algorithm includes robustness towards noise, articulated variations, and rigid transformation, speed (no need of GPU), application in real-time scenarios; Improvement to DTW algorithm by using Chi-Square coefficient.

It was found that the proposed algorithm outperformed many at the time existing algorithms on different datasets and results are shown in Table 1 (Zhu et al., 2021).

Table 1: 3D shape context - Performance on datasets

| Dataset | Accuracy (%) |
| --- | --- |
| NTU Hand Digit Dataset | 98.7 |
| Kinect Leap Dataset | 96.8 |
| Senz3d Dataset | 99.6 |
| ASL-FS Dataset | 87.1 |
| ChaLearn LAP IsoGD Dataset | 60.12 |

Despite performing well on many datasets, it was found that the accuracy was not good enough on many other datasets and it can be concluded that the approach though outperformed many other algorithms was not good enough to be generalized and will not be the best option in real-time applications.

A new experiment was done, utilizing NAO robot[22][23][27][28][29][30] to evaluate the effectiveness of the model. The said technique used Leap Motion (Wu et al., 2022) to gather data and applied Kalman filter to the original data to remove unwanted noise. From the original coordinates there were three new features that were extracted and used namely, angle feature, angle velocity feature and length feature. Finally, these extracted features were fed to LSTM-RNN network to predict the gesture. Below Figure 3 (Wu et al., 2022) is the pipeline demonstrating the overall architecture.

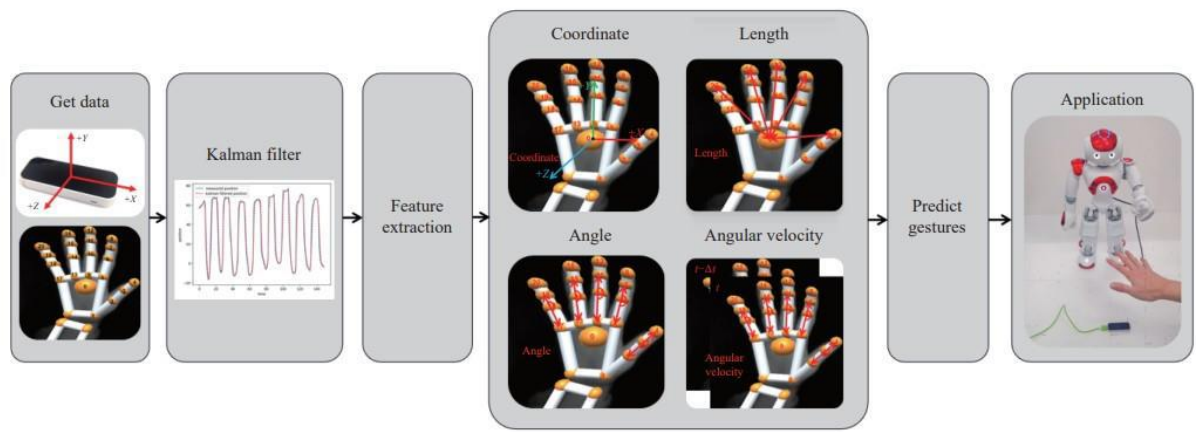

Figure 3: LSTM-RNN network pipeline

Several experiments were done to find the accuracy of the model and it was found that using just 3D positions of finger joints resulted in 97.93% accuracy, only length resulted in 95.17%, Angle in 93.79% and Angular Velocity resulted in very less accuracy i.e. 79.31% due to speed the player used in movements. However, when all features were used together, highest accuracy of 99.31% was achieved.

Though the accuracy achieved in the research is very good but needs to be verified on other datasets and is lacking in the current research. Another aspect is noise introduced in Leap Motion and is something that can be studies further and instead of Kalman filter, other filters can be tried, and overall impact can be studied.

**Graph based Gesture Recognition**

In above approaches we saw how an image can be used to extract feature and deep learning techniques like LSTM etc can be applied. Feature creation is one of the most important facts and a good feature can help train a model and achieve a very good accuracy. Skeleton data now a days is widely used due to their robustness to accommodate complex and dynamic circumstances in action recognition. Along with feature creation and trying different datasets and data creation techniques, new techniques need to be evolved and one of such technique was use of Graph based techniques. In this section we will study different graph-based techniques that have been applied and pros and cons of same. Conventional methods utilizing skeleton data relied heavily on hand crafted features to extract skeleton information, however with development of Deep Learning methods like CNN, RNN and GCN's, more advance mechanism was derived.

Earlier, joint information was used from skeleton data and temporal analysis was done for action recognition, however since they were lacking utilization of spatial information which is crucial in action determination, a new methodology ST-GCN (Yan et al., 2018) was developed which uses both spatial and temporal information to identify the action. Below Figure 4 (Yan et al., 2018) shows graph capturing spatial-temporal information.

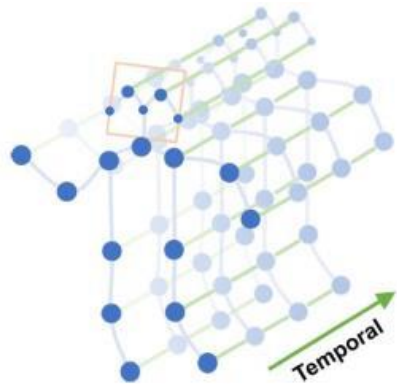

Figure 4: The spatial temporal graph of a skeleton sequence

From the figure, we can infer that there are two types of edges, one form basis of spatial i.e.. Natural connectivity of joins and other that connects joins within different time frame and called as temporal edge. Below Figure 5 (Yan et al., 2018) visually describes the method used in ST-GCN

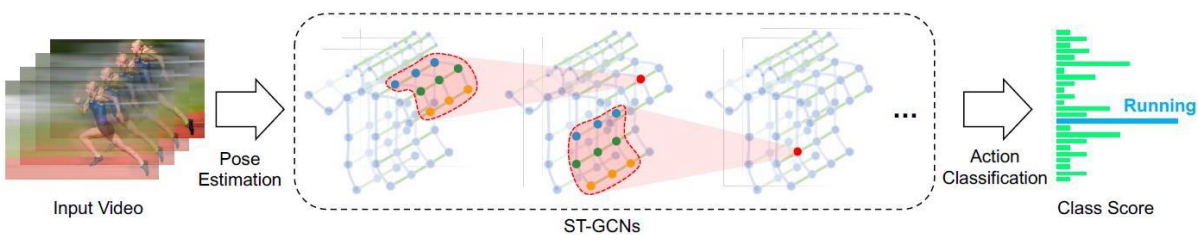

Figure 5: Construct spatial temporal graph on skeleton sequences

ST-GCN was evaluated over two datasets namely Kinetics and NTU-RGB+D, ST-GCN reached accuracy of 52.8% and 88.3% respectively. The technique opened doors to experiments using GCN and further studies were done on how to create feature from skeleton data.

Though ST-GCN proved to be promising method, but it was found that node interaction does not necessarily provide the complementary required information, it also introduces possibility of noise. It was also found that use of GCN can become over-smoothing when multi-layer GCN is used.

To overcome the issues of GCN, a new methodology ST-GDN (Peng et al., 2021) was proposed. This method provides a better aggregation of messages by removing embedding redundancy, it addresses GCN's over-smoothing problem. High level working of this method is shown in Figure 6 (Peng et al., 2021)

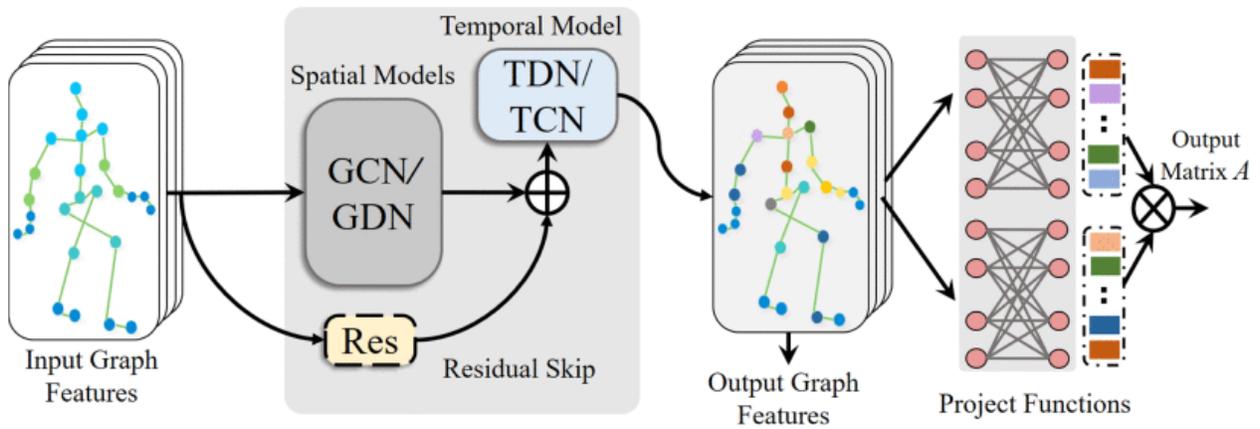

Figure 6: Illustration of the ST-GDNs block

ST-GDNs basically has four building blocks, listed in table below with high level working.

Table 2: ST-GDNs building blocks

| | |
|---|---|
| Node-wise ST-GDN (ST-GDN2) | Represents the features in new feature-space (coordinates changed), the feature embeddings are standardized and corelation is removed, thereby removing the over-smoothing problem |
| frame-wise ST-GDN (ST-GDN-T) | Like ST-GDN2 |
| element-wise ST-GDN (ST-GDN-E) | Like ST-GDN2 |
| combination of GCN and GDN (ST-GDCN) | Graph representation learning is enhanced by use to of convolutional/deconvolutional feature embeddings |

Below Figure 7 (Yan et al., 2018; Peng et al., 2021) shows feature creation by both ST-GCN and ST-GDN methods.

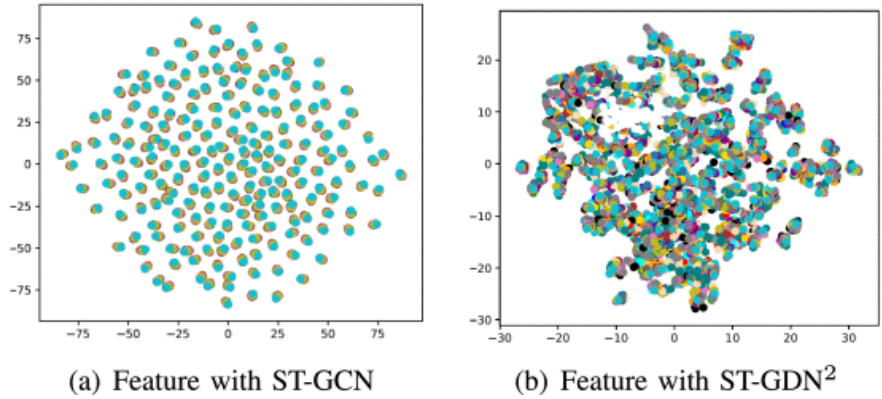

(a) Feature with ST-GCN  (b) Feature with ST-GDN$^2$

Figure 7: Feature creation

As we can see that in Figure 7: (a) ((Yan et al., 2018; Peng et al., 2021) we find that it is hard to distinguish nodes as they have similar representation, on the contrary when coordinates are changed, nodes can be easily distinguished hence confirming ST-GDN2, alleviate over-smoothing problem.

Above method was evaluation over many challenging datasets and listed are accuracies against each.
- NTU RGB+D dataset – 95.9%
- NTU RGB+D 120 dataset – 82.3%
- Kinetics-skeleton dataset – 60.5%

Although, ST-GCN were successful in hand gesture recognition, it was found that they had certain limitation like usage of fixed graph be it spatial based on hand skeleton tree or temporal dimension which restricts hand gesture recognition. Actions like touching forehead is different from clapping or jumping, these examples strongly indicates that graph structure should be data dependent and this problem is not solved in ST-GCN. In order to overcome these issues two-stream graph attention convolutional network with spatial–temporal attention was proposed (Zhang et al., 2020) which was based on (Wang et al., 2022).

Beginning with Adaptive Graph Convolutional Network (Wang et al., 2022), it basically constructs two types of graphs, one can be called as global graph which represents common data patterns and other is unique to local(each) data points. These two graphs are then optimized individually which helps in better fitting of the model's hierarchical structure. ST-GCN focusses on first order information which is feature vector of vertex, but it does not consider second order information which is basically feature of bones between joints. This information is crucial as bone length and direction plays an important role in action recognition, AGCN on other hand utilizes this information. AGCN formulates length of bones and their direction as vector, this vector is then fed to AGCN to predict the class. Basic idea of AGCN is to utilize both networks and increase the efficiency.

The spatiotemporal graph convolution is based on predefined graph and as described above it has limitations, AGCN solves this problem by formulating a way to optimize all other parameters together and forms an end-to-end learning. It basically works on connection between two vertexes and their connectivity strength. Below Figure 8 (Wang et al., 2022) shows overall architecture of 2 Stream AGCN, where B-Stream stands for network of bones and J-Stream stands for network of joints. From both J and B stream we get scores which are finally added and fed to SoftMax layer for prediction.

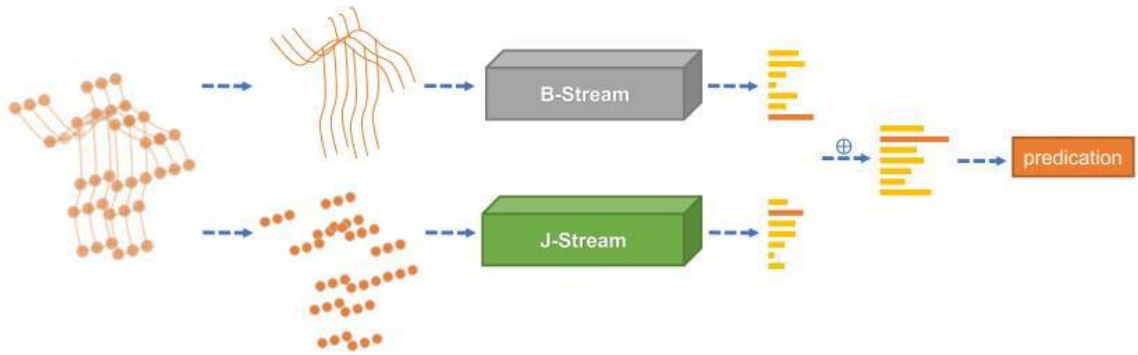
Figure 8: 2S-AGCN architecture

The two stream – AGCN method was evaluated over two datasets. It was found that the model showed 95.1% accuracy over NTU-RGBD and 58.7% on Kinetics-Skeleton dataset.
Two stream- AGCN influenced two-stream graph attention convolutional network with spatial–temporal attention (STA-GCN) (Zhang et al., 2020), basic idea was to use motion over bone stream and better results were achieved.
STA-GCN, in addition to utilizing concepts of two stream – AGCN, it can be summarized in two steps as below:

- Temporal graph attention module was used, so that hand gesture encoding can be done with multi-scale temporal features.

- Two-stream hand gesture network was used as briefed below:

    o Pose stream – it uses joints from each frame as input

    o Motion stream – it uses joint offsets between neighbouring frames

Below Figure 9 (Zhang et al., 2020) details out network architecture for single stream.

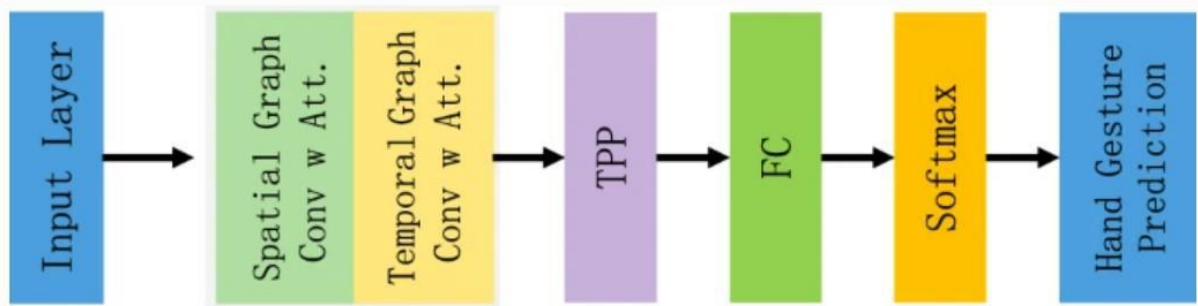
Figure 9: Single Stream architecture

As shown in above figure, is the network architecture used by both Pose and Motion streams. First, we initialize the skeleton graph, the input then is fed to Spatial Graph convolution with spatial graph attention mechanism and temporal graph attention to extract the spatial temporal features.
Figure 10 (Zhang et al., 2020) shows spatial temporal features, where black line denotes spatial connections and blue denotes temporal connections.

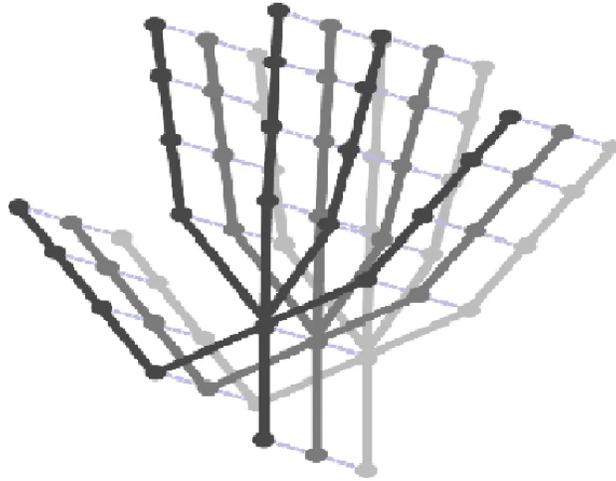

Figure 10: The spatial–temporal joint connections of the initial graph

The output feature of GCN with spatial graph attention will be fed to the GCN with temporal graph attention. Temporal pyramid pooling layer (TPP) (Wang et al., 2017), is then used to extract multi-scale features, the output of TPP is then fed to fully connected (FC) layer and SoftMax is used for classification.

Below Table 3 (Zhang et al., 2020) shows accuracy of STA-GCN when used of SHREC'17

Table 3: STA-GCN accuracy matrix

| Stream | 14 gestures |
|---|---|
| Pose stream | 93.2 |
| Motion stream | 94.4 |
| Two streams | **94.5** |

On DHG14/28 dataset, on 14 gestures accuracy of 91.5% was achieved.

STA-GCN when compared with 2s-AGCN on SHREC'17, it was found to be better by 2.1% on 14 gestures recognition and 0.7% on 28 gestures recognition, while when compared on DHG14/28, STA-GCN outperformed 2s-AGCN by 1.6% on 14 gestures recognition and 1.2% on 28 gestures recognition.

In 2019 (Shi et al., 2019), another improvement to graph-based methods was done, earlier methods used bones and joints separately and hence making skeleton as undirected graph, this posed limitations. To overcome these limitations, skeleton was representing as DAG, joints were used as vertexes and bones represents edges. Using this information, a DGCN was designed, which could propagate the information in adjacent joints and bones and hence better represent the current state, thereby giving opportunity to better identify the action. This study also solves the problem where graph cannot be directly used to populate the coordinates, for example clapping or hugging. Since there is strong dependency between two body parts be it two hands or hugging someone, such feature cannot be constructed using graph. This problem was solved using adaptive graph, in which topology of graph is parameterized and optimization takes place during training. Below Figure 11 (Shi et al., 2019) is graph representation of human body, where blue circle indicates the root vertex.

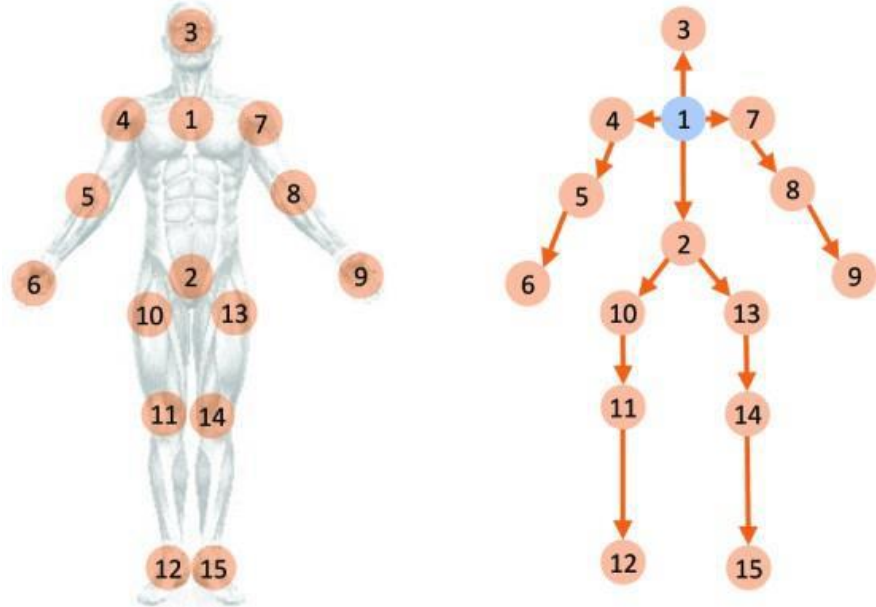

Figure 11: Illustration of the graph construction for skeleton data

Following are the high-level steps involved:

• Prepare Skeleton data frames containing joint coordinates

• Extract bone information

• Represent Spatial information of joint and bones as vertexes and edges within a DAG

• DGCN – Extract the action recognition features

DGCN method was applied on two famous datasets NTU-RGBD and Skeleton-Kinetics and accuracy of 96.1% and 59.6% was achieve respectively. Main thing to note here is due to diversity of actions in Skeleton-Kinetics dataset we find that accuracy is not very good, and this indicates that more study in the field is required and methods which can be generalized with higher accuracy.
This study paves a path where generalization is key, use of skeleton + RGB data together can be studied and possibly a better accuracy can be achieved.
**Point Based Gesture Recognition**
Above section, we learnt use of Graph based methodologies in gesture recognition, idea was very clear and was to make use of joints in human body and derive methods which are efficient. Different other techniques were then applied on graph to make the recognition generic.

In 2010 (Li et al., 2010), 3D points were used from the depth maps to classify the action. During this time there was no public dataset (benchmark) available and hence a dataset having 20 actions was collected for study. In total 7 subjects performed each action 3 times to create a dataset of 4020 action samples with depth map of 640X480. The depth image was first down sampled to reduce computational cost, the sampled 3D points were then allocated to XY, YZ and XZ projections. For all conducted experiments, , training samples were clustered using the Non-Euclidean Relational Fuzzy (NERF) C-Means and the dissimilarity between two depth maps was calculated as the Harsdorf distance between the two sets of the sampled 3D points as shown in Figure 12 (Li et al., 2010) below.

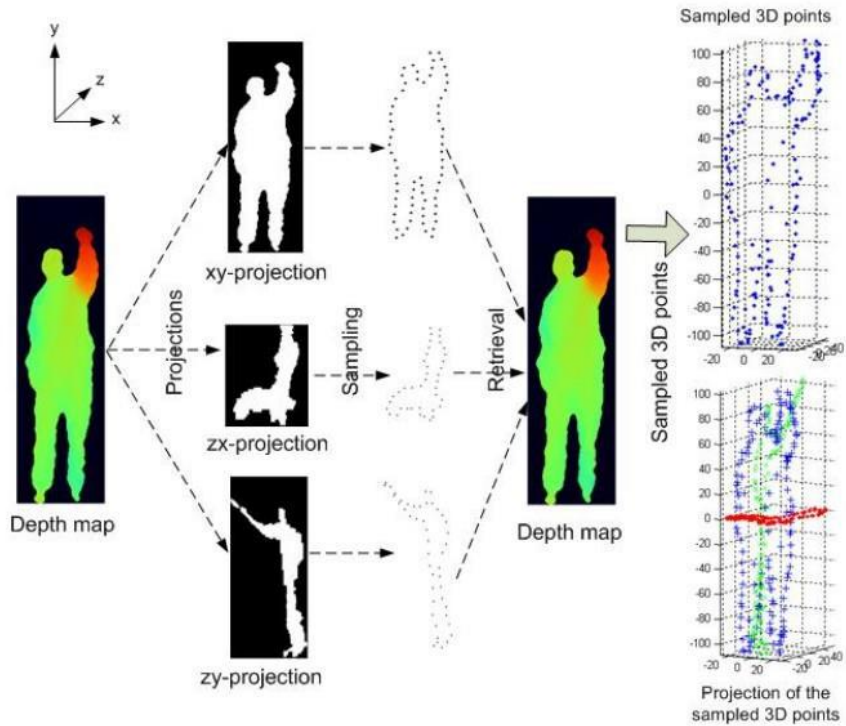

Figure 12: Bag of 3D points from depth image

To evaluate the methodology, the datasets were divided into 3 parts, details of datasets and accuracy are listed in Table 4 below.

Table 4: NERF based accuracy matrix

| Train-Test split – 1/3-2/3 | | |
|---|---|---|
| AS1 | Simple action of one kind | 89.5% |
| AS2 | Simple action of another kind | 89.0% |
| AS3 | Complex actions | 96.3% |
| Overall | | 91.6% |

This study also observed differences in accuracies when different quadrant of images was removed (images was divided in 4 quadrants), it was concluded that part which had least significance is removed then accuracies increases and vice-versa, this study hence paved a path to concentrate on points which are significant for that recognition. It was found that 3D points offer a lot, and more research were required to get to better and generic methods.

2014 (Vemulapalli et al., 2014), a new methodology which represented action as a curve in the lie group was proposed. As there are difficulties in classifying the curves in lie group, mapping of curves in lie group was done to the lie algebra, which is nothing but a vector space and hence easy to model for classification. The vectors were then used with DTW, Fourier temporal pyramid representation and linear SVM. Below Figure 13 (Vemulapalli et al., 2014) shows curve in the lie group and Figure 14 (Vemulapalli et al., 2014) shows methodology used.

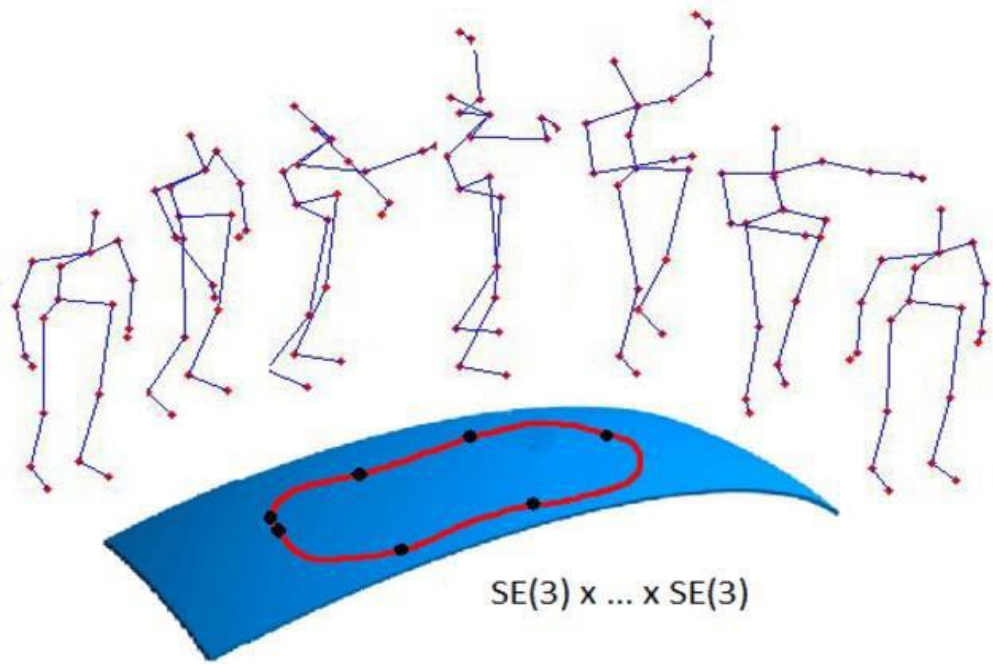

Figure 13: Representation of an action (skeletal sequence) as a curve in the Lie group SE(3) ×. ..

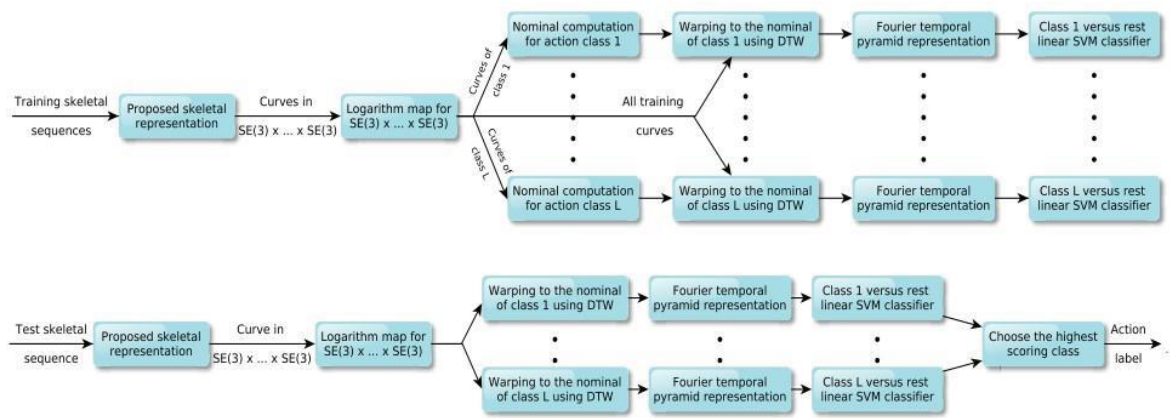

Figure 14: Training and Testing pipeline using Lie Group

The model was evaluated on many datasets and found to be effective, below Table 5 shows the accuracies.

Table 5: Lie group-based accuracy matrix

| Datasets | Accuracies |
| --- | --- |
| MSR-Action3D | 92.46 |
| UTKinect | 97.08 |
| Florence3D | 90.88 |

Though the model results were good, but usage was limited due to following reasons:
• Dynamic identification of body part used for action recognition was unavailable

• Actions performed were by a single person and hence accuracies cannot be generalized.

Many research were being done during these years, focused on cloud based points, taking the studies further, in 2018 (Ge et al., 2018), Hand PointNet was proposed. This methodology was different from other CNN based (methods using CNN had issue containing the space and time complexities which can grow cubically) from the fact that this method directly processed the 3D point cloud which was formed from the visible hand surface for pose regression. To improve the efficiency of the model, neighboring finger points were also considered and used in the modelling and referred to as fingertip refinement network. Below Figure 15 details out the architecture used in Hand PointNet.

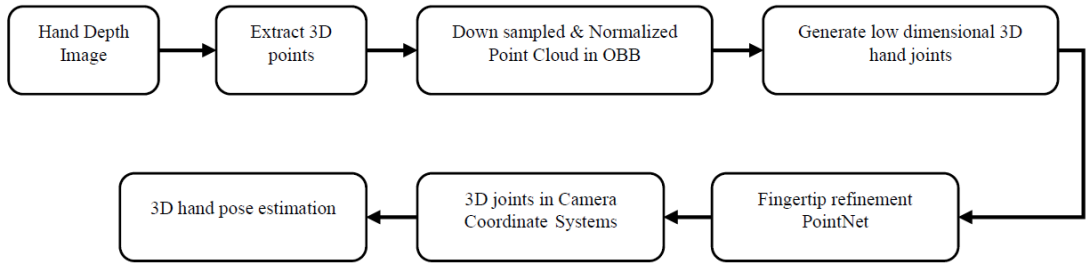

Figure 15 Hand PointNet architecture

OBB – It helps solve the large variation in hand orientation, by transforming the original hand point cloud into a canonical coordinate, making global orientation consistent.
Fingertip refinement PointNet – Takes k-nearest neighboring points of fingertip location and outputs refined 3D fingertip locations, with this we can

• Decrease fingertip estimation error

• KNN will not change even if there is deviation of fingertip location from ground truth.

The methodology was compared against many other methods, and it was found that PointNet was superior to other as shown in Figure 16 (Ge et al., 2018) below.

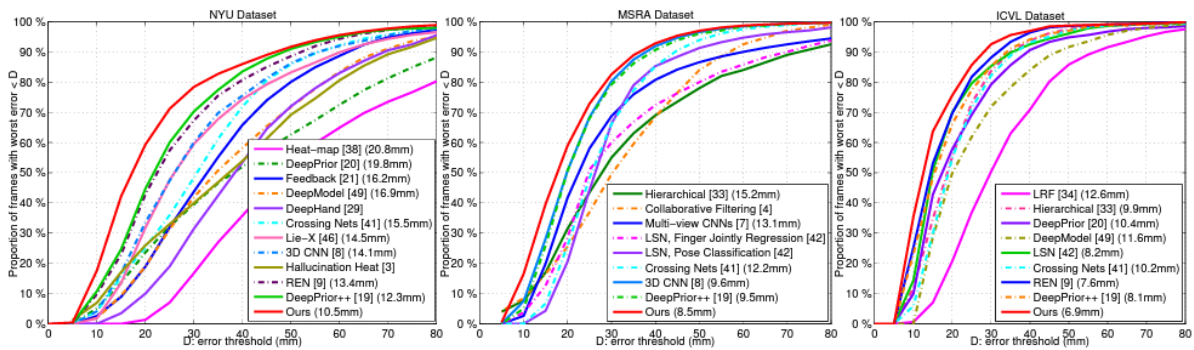

Figure 16: Comparison with other techniques (Our stands for PointNet)

Point cloud at this stage has become one of most studied and many methodologies were proposed. In continuation on how point clouds can further be utilized, in 2019 (Liu et al., 2019b), Meteornet was proposed. Specialty of this proposed method was that it directly consumes the dynamic sequences of point clouds and learns both local and global features and this learning was utilized to solve problems like classification, segmentation etc.

MeteorNet is based on a novel neural network module called Meteor module, specialty of Meteor is that it takes point clouds and learn features of each point from it by aggregating the spatiotemporal neighborhoods. Design is such that it can learn from previous features by stacking modules over one another. Finally, the stacked modules can capture information from larger neighborhood. However, problem with this method was its inability to properly determine the spatiotemporal neighborhoods while the object is in motion. To address this, under MeteorNet two methods were proposed, shown in Figure 17 (Liu et al., 2019b):
• Direct grouping – Directly increases the grouping radius over time

• Chained grouping – Tracks the motion of the object and uses offline estimated scene flow to construct the neighbourhood.

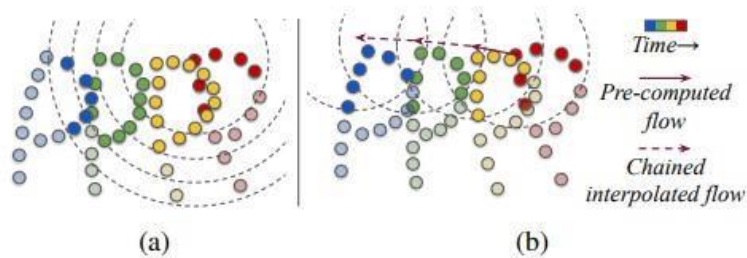
Figure 17: (a) direct grouping; (b) chained-flow grouping

MeteorNet evaluation was done on many datasets, and below Table 2.6 shows the accuracy achieved on classification dataset (limited due to scope of this study).

Table 7: Classification accuracy on MSRAction3D (%)

| # Of Frames | Accuracy |
|---|---|
| 4 | 78.11 |
| 8 | 81.14 |
| 12 | 86.53 |
| 16 | 88.21 |
| 24 | 88.50 |

Another methodology, FlowNet3D (Liu et al., 2019a) based on deep neural network was proposed which is capable of learning flow of scene using point clouds in an end-to-end fashion. The architecture is such that it can learn hierarchical feature from point clouds and flow embeddings representing point motions simultaneously.

FlowNet3D utilizes two consecutive flow frames let's call then point cloud 1 and point cloud 2, the network inside FlowNet3D is capable of estimating a translation flow vector for every single point in point cloud 1 (with help of point cloud 2) thereby indicating the motion(flow) between two frames as shown in Figure 18 (Liu et al., 2019a).

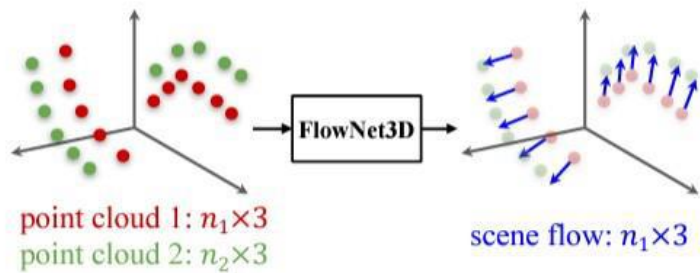

Figure 18: End-to-end scene flow estimation from point clouds

FlowNet3D has 3 building blocks as explained below on overall how it works.

• Point Feature Learning – Since point clouds are irregular and order less, traditional convolutions are not effective and hence a new architecture PointNet++ (Qi et al., 2017) was used which is capable of learning hierarchical features.

• Point Mixture – Mixing of two-point clouds is not an easy task due to viewpoint shift and occlusions. Idea was based on finding displacement between two points at time t and t+1 and using weights to achieve the end state.

• Flow Refinement – Here the flow embedding found using Point Mixture are up-sampled to the original points and helps in propagating the points features so that they can be learnt.

To evaluate the model, end point error (EPE = average distance between estimated flow vector and ground truth) and flow estimation accuracy (ACC = portion of estimated flow vector which are below EPE threshold) was used.
Results when evaluated on FlyingThings3D were as follows:

• EPE = 0.1694

• ACC (<5%) = 25.37%

• ACC (<10%) = 57.85%

There were multiple research using point cloud and various different methods were applied, another such techniques which utilizes point cloud based on LSTM was proposed in 2020 (Min et al., 2020). In this methodology idea was to propagate the information from past to future while retaining the spatial structure. This method combines past state information of neighboring points with current state and with the help of weight-shared LSTM. One of problems of all methods described was that they were unable to capture long term relationships and were basically short term. Ideally a point at time t-1 will have representation at time t but, it won't be easy to find the exact point and that is where this method has been useful. It used the neighboring points to find the state.
PointLSTM was found effective on datasets like SHREC'17 and NVGestures. On SHREC'17, on 14 gesture it reported max of 95.9 % and on 28 gestures, 94.7%. On NVGesture it was found to be 87.9% accurate.

Another LSTM based approach, Two Stream LSTM (Gammulle et al., 2017) was proposed which focused on learning salient spatial features using CNN and finally using LSTM to map the temporal relationship. Proposed techniques showed that the output of both layers combined has greater recognition ability than using either stream alone. The results showed that a fully connected layer output can be used as a tool to direct the LSTM to the important parts of the convolutional feature sequence. The proposed mechanism achieved 94.6 %, 99.1%, 69.0 % accuracy on UCF Sport, UCF11 and jHMDB datasets respectively.

Another study which also based on the fact that the single-stream models are not adequate for capturing both fine-grained local posture variations and global hand movements was proposed (Bigalke and Heinrich, 2022). The idea was to decouple learning of local and global features using a dual-stream model, the features were then fused to a LSTM layer where temporal learning was done. Through leveraging the complementary benefits of raw point clouds and BPS-based representations, proposed framework explicitly learns global position and local posture features as shown in Figure 19 (Bigalke and Heinrich, 2022). The approach was computationally more efficient, but the accuracy could have been increased by adding an extra module for low-level spatiotemporal feature extraction.

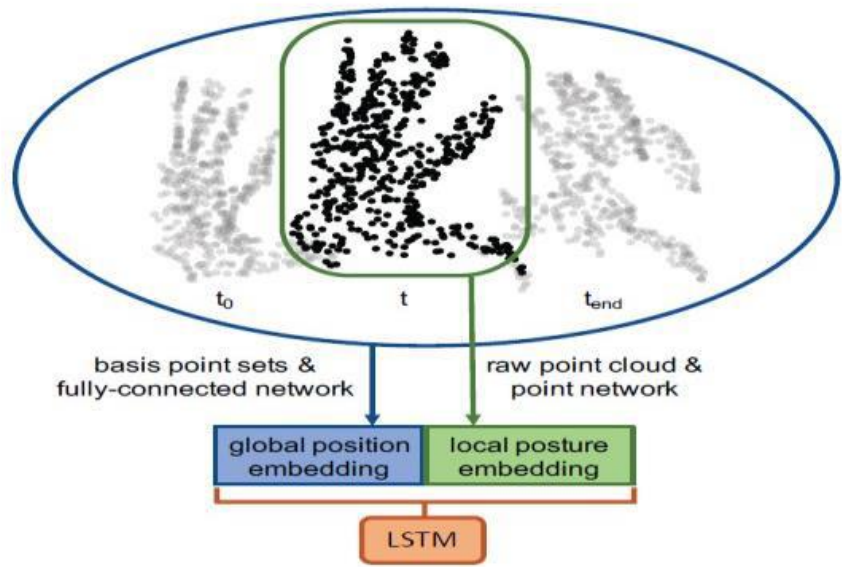

Figure 19: 2S model to capture both global position and local posture features

The method was found to be highly effective on datasets like SHREC'17 and achieved 96.1 and 95.2 % accuracy on 14 and 28 Gestures respectively.

**Point Clouds nearest neighbors and sampling**
Efficiently finding nearest neighbor in point cloud is an important task, as we have studies in other research papers that information about a point can be derived using its neighbors, however more the effective method is less the noise that will get propagated from one frame to another. Another aspect is to sample the 3D point clouds, as the data is huge we also need to make sure that we do not oversample and same time simplify the 3D point cloud data.

Representing the point clouds is an important task and a study was done in 2010, (Wang et al., 2010) and a simplified way to represent 3D point clouds were proposed. Idea was to integrate both the feature parameters and unform spherical sampling. The first thing we do is establish parameters comprised of the average of neighboring point's average distance, angle formed between point and neighboring points and point curvature, another step is then is to define the feature threshold by calculating density of 3D points, this helps in differentiating between feature and non-feature points. Uniform sampling is then applied over the non-feature points. Figure 20 (Wang et al., 2010) shows feature and non-feature points. Spherical parameterization techniques can be used for sampling.

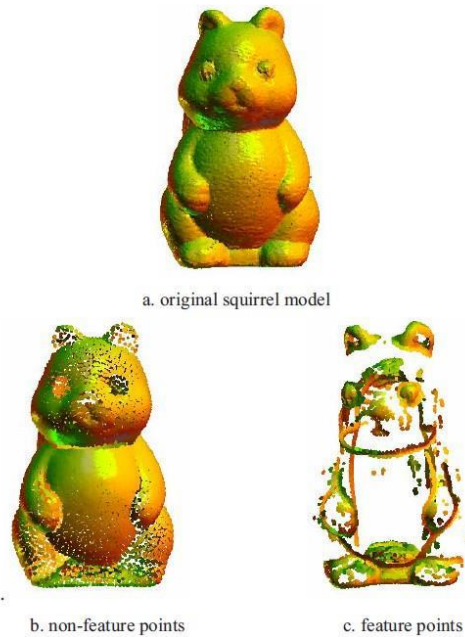

Figure 20: Feature and Non-Feature points

The benefit achieved by this method was that it shows ways to reduce the 3D point cloud data and same time retaining the sharp information without loss.

Continuing with the previous studies on how to effectively create optimized 3D points was proposed in 2011 using k nearest neighbor (Angelo and Giaccari, 2011). 3D point clouds need to be processed so as to construct high level information beginning from cartesian coordinates. The process of processing point clouds usually entails removing residual noise, adjusting the sampling rate, estimating the points' normal, and/or tessellation. Various operations in the cloud require computing the k-nearest neighborhoods (KNN) of each point. Angelo and Giaccari proposed knn search based on new data structure using typical space division approach. It was comprised of two basic steps:
1. Data Structure creation – Helps reducing computational cost
2. Nearest-neighbour search

With success in using KNN on point clouds another approach, an improved method based on dimension reduction and sorting was proposed for the K-nearest neighbors algorithm (Li and Wang, 2017). PCA was introduced to analyze the spatial distribution of point clouds data main directions. In the next step, turn the main directions so they match the X, Y, and Z coordinates, sort the point cloud data in the three coordinate axes, and find the position of the query point. Next the distance between the query point and its neighbors by extracting neighbor points in proportion to the three sorted point cloud data sets. In order to determine the nearest neighbors, sort the distance by the first k points found. The proposed algorithm helped in reducing point to point distance calculation time efficiently. The KNN algorithm was improved, resulting in a major time savings when solving point cloud normal vectors, reconstructing surfaces, and similar operations. Exploring feature of neighboring point was possible based on studies done, however the existing methods suffered from ambiguous feature, mainly junction regional points. An approach using segmentation scores of neighboring points was proposed in 2019 (Zhao et al., 2019). The module proposes a method for improving 3D point cloud segmentation scores by utilizing attention-based refinement, which can be easily integrated with existing networks. This study allowed possible to focus on a particular region to extract required feature which can be used for modeling.

Table 7: Summary of different methods used in Gesture recognition

| | Technique | Advantages | Disadvantages | Achievement |
|---|---|---|---|---|
| Vision Based | 3D shape context was used to represent 3D hand gestures | robustness towards noise, articulated variations, and rigid transformation, speed (no need of GPU), application in real-time scenarios | Approach cannot be generalized and not good on real time scenarios | Improvement to DTW algorithm by using Chi-Square coefficient |
| Vision Based | Hand joint coordinate features collected by the Leap Motion and training using LSTM-RNN | Application in robotics and high accuracy | Noise introduced in leap motion needs to be studied further | Methodology can be used extensively in robotics |
| Graph based | ST-GCN - Spatial-temporal graph convolution network- uses both spatial and temporal information to identify the action | Made GCN popular and creation of features from skeleton data | Node interaction does not necessarily provide the complementary required information, it also introduces possibility of noise. It was also found that use of GCN can become over-smoothing when multi-layer GCN is used | Good accuracy achieved over NTU-RGB+D |
| Graph based | ST-GDN - Spatial Temporal Graph Deconvolutional Network for Skeleton-Based Human Action Recognition | Better aggregation of messages by removing embedding redundancy, it addresses GCN's over-smoothing problem | limitation like usage of fixed graph be it spatial based on hand skeleton tree or temporal dimension which restricts hand gesture recognition | Improvement over ST-GCN, good accuracy on • NTU RGB+D and • NTU RGB+D +120 datasets |
| | 2S AGCN and STA-GCN Two-stream graph attention convolutional network with spatial–temporal attention | Overcame ST-GDN limitations and made graph structure data dependent hence classifying actions like forehead touching | Making use of joints and bones makes skeleton graph undirected, hence state presentations suffered some limitations | Temporal graph attention module was used, so that hand gesture encoding can be done with multi-scale temporal features Achieved higher accuracy on NTU-RGBD datasets |
| | DGCN | Skeleton was representing as DAG, joints were used as vertexes and bones represents edges; better represent the current state, thereby giving opportunity to better identify the action | Diversity of actions impacts the accuracy, and hence more efforts are requried to generalize the classification of gestures | Solves the problem where graph cannot be directly used to populate the coordinates, for example clapping or hugging; High accuracy on NTU-RGBD |
| Point based | Bag of 3D points from depth image | Benchmark for study in point clouds and data collection | Limited scope and data were collected for this study itself; limited accuracy | Focus on action area increased the accuracy of the model. It was found that 3D points offers a lot and more researches were required to get to better and generic methods |
| | Methodology which represented action as a curve in the lie group | Good accuracy on datasets like MSR-Action3D | Dynamic identification of body part used for action recognition was unavailable Actions performed were by a single person and hence accuracies cannot be generalized. | New way of representation and option for further studies |
| | Hand PointNet - 3D Hand Pose Estimation using Point Sets | Overcome issue using CNN; CNN had issue containing the space and time complexities which can grow cubically | | Directly processed the 3D point cloud which was formed from the visible hand surface for pose regression; as fingertip refinement network |
| | MeteorNet: Deep Learning on Dynamic 3D Point Cloud Sequences | Can learn from previous features by stacking modules over one another; the stacked modules are able to capture information from larger neighborhood | Point clouds are too sparse, and the flow estimation is inaccurate. In this case, errors may accumulate during chaining, and the resulting spatiotemporal neighborhood may deviate from the true corresponding | directly consumes the dynamic sequences of point clouds and learns both local and global features and this learning was utilized to solve problems like classification, segmentation etc. |

**Notes:** For more information about different applications of AI and deep learning techniques please go through these papers [6] [7] [8] [9] [10] [11] [12] [13] [14] [15] [16] [17] [18] [19] [20] [21] [22] [23] [24] [25] [26] [27] [31] [32] [33].

**References**
Angelo, L. Di and Giaccari, L., (2011) An efficient algorithm for the nearest neighbourhood search for point clouds. *International Journal of Computer Science Issues*, 85, pp.1–11.
Apostol, B., Mihalache, C.R. and Manta, V., (2014) Using spin images for hand gesture recognition in 3D point clouds. *2014 18th International Conference on System Theory, Control and Computing, ICSTCC 2014*, November 2010, pp.544–549.
Bigalke, A. and Heinrich, M.P., (2022) Fusing Posture and Position Representations for Point Cloud-Based Hand Gesture Recognition. pp.617–626.
Celebi, S., Aydin, A.S., Temiz, T.T. and Arici, T., (2013) Gesture recognition using skeleton data with weighted dynamic time warping. *VISAPP 2013 - Proceedings of the International Conference on Computer Vision Theory and Applications*, 1, pp.620–625.
Chao, Z., Pu, F., Yin, Y., Han, B. and Chen, X., (2018) Research on real-time local rainfall prediction based on MEMS sensors. *Journal of Sensors*, 2018, pp.1–9.
Du, Y., Fu, Y. and Wang, L., (2015) Skeleton Based Action Recognition with Convolutional Neural Network Nat ' l Lab of Pattern Recognition , Institute of Automation , Chinese Academy of Sciences. *2015 3rd IAPR Asian Conference on Pattern Recognition*, pp.579–583.
Epfl, D.L., (2018) POINT CLOUD QUALITY ASSESSMENT METRIC BASED ON ANGULAR SIMILARITY Evangelos Alexiou and Touradj Ebrahimi Multimedia Signal Processing Group ( MMSPG ) Ecole Polytechnique F ´ Emails : FirstName.LastName@epfl.ch. *2018 IEEE International Conference on Multimedia and Expo (ICME)*, pp.1–6.
Gammulle, H., Denman, S., Sridharan, S. and Fookes, C., (2017) Two stream LSTM: A deep fusion framework for human action recognition. *Proceedings - 2017 IEEE Winter Conference on Applications of Computer Vision, WACV 2017*, pp.177–186.
Ge, L., Cai, Y., Weng, J. and Yuan, J., (2018) Hand PointNet: 3D Hand Pose Estimation Using Point Sets. *Proceedings of the IEEE Computer Society Conference on Computer Vision and Pattern Recognition*, pp.8417–8426.
Hardin, P.J., (1999) Comparing main diagonal entries in normalized confusion matrices: a bootstrapping approach. *International Geoscience and Remote Sensing Symposium (IGARSS)*, 1801, pp.345–347.
Hoang, N.N., Lee, G.S., Kim, S.H. and Yang, H.J., (2019) Continuous Hand Gesture Spotting and Classification Using 3D Finger Joints Information. *Proceedings - International Conference on Image Processing, ICIP*, 2019-Septe, pp.539–543.
Ikram, A. and Liu, Y., (2020) Skeleton based dynamic hand gesture recognition using LSTM and CNN. *ACM International Conference Proceeding Series*, pp.63–68.
Li, D. and Wang, A., (2017) Improved KNN algorithm for scattered point cloud. *Proceedings of 2017 IEEE 2nd Advanced Information Technology, Electronic and Automation Control Conference, IAEAC 2017*, pp.1865–1869.
Li, W., Zhang, Z. and Liu, Z., (2010) Action recognition based on a bag of 3D points. *2010 IEEE Computer Society Conference on Computer Vision and Pattern Recognition - Workshops, CVPRW 2010*, pp.9–14.
Li, Y., He, Z., Ye, X., He, Z. and Han, K., (2019) Spatial temporal graph convolutional networks for skeleton-based dynamic hand gesture recognition. *Eurasip Journal on Image and Video Processing*, 20191.
Lin, H., Hsu, M. and Chen, W., (2014) Human Hand Gesture Recognition Using a Convolution Neural Network. pp.1038–1043.

Lin, X., Sánchez-Escobedo, D., Casas, J.R. and Pardàs, M., (2019) Depth estimation and semantic segmentation from a single RGB image using a hybrid convolutional neural network. *Sensors (Switzerland)*, 198. 106

Liu, X., Qi, C.R. and Guibas, L.J., (2019a) Flownet3d: Learning scene flow in 3D point clouds. *Proceedings of the IEEE Computer Society Conference on Computer Vision and Pattern Recognition*, 2019-June, pp.529–537.
Liu, X., Yan, M. and Bohg, J., (2019b) Meteornet: Deep learning on dynamic 3D point cloud sequences. *Proceedings of the IEEE International Conference on Computer Vision*, 2019-OctobIccv, pp.9245–9254.
Mack, Y.P. and Rosenblatt, M., (1979) Multivariate k-nearest neighbor density estimates. *Journal of Multivariate Analysis*, 91, pp.1–15.
Min, Y., Zhang, Y., Chai, X. and Chen, X., (2020) An Efficient PointLSTM for Point Clouds Based Gesture Recognition. *Proceedings of the IEEE Computer Society Conference on Computer Vision and Pattern Recognition*, pp.5760–5769.
Nguyen, X.S., Brun, L., Lezoray, O. and Bougleux, S., (2019a) A neural network based on spd manifold learning for skeleton-based hand gesture recognition. *Proceedings of the IEEE Computer Society Conference on Computer Vision and Pattern Recognition*, 2019-June, pp.12028–12037.
Nguyen, X.S., Brun, L., Lézoray, O. and Bougleux, S., (2019b) Skeleton-based hand gesture recognition by learning SPD matrices with neural networks. *Proceedings - 14th IEEE International Conference on Automatic Face and Gesture Recognition, FG 2019*.
Nishida, N. and Nakayama, H., (2016) Multimodal gesture recognition using multi-stream recurrent neural network. *Lecture Notes in Computer Science (including subseries Lecture Notes in Artificial Intelligence and Lecture Notes in Bioinformatics)*, 9431, pp.682–694.
Peng, W., Shi, J. and Zhao, G., (2021) Spatial Temporal Graph Deconvolutional Network for Skeleton-Based Human Action Recognition. *IEEE Signal Processing Letters*, 28, pp.244–248.
Qi, C.R., Yi, L., Su, H. and Guibas, L.J., (2017) PointNet++: Deep hierarchical feature learning on point sets in a metric space. *Advances in Neural Information Processing Systems*, 2017-Decem, pp.5100–5109.
Shi, L., Zhang, Y., Cheng, J. and Lu, H., (2019) Skeleton-based action recognition with directed graph neural networks. *Proceedings of the IEEE Computer Society Conference on Computer Vision and Pattern Recognition*, 2019-June, pp.7904–7913.
Shih, H.C. and Ma, C.H., (2018) Hand gesture recognition using color-depth association for smart home. *Proceedings - 2018 1st International Cognitive Cities Conference, IC3 2018*, pp.195–197.
Sutskever, I., Vinyals, O. and Le, Q. V., (2014) Sequence to sequence learning with neural networks. *Advances in Neural Information Processing Systems*, 4January, pp.3104–3112.
Vedula, S., Baker, S., Rander, P., Collins, R. and Kanade, T., (2005) Three-dimensional scene flow. *IEEE Transactions on Pattern Analysis and Machine Intelligence*, 273, pp.475–480.
Vemulapalli, R., Arrate, F. and Chellappa, R., (2014) Human action recognition by representing 3D skeletons as points in a lie group. *Proceedings of the IEEE Computer Society Conference on Computer Vision and Pattern Recognition*, pp.588–595.
Wang, L., Chen, J. and Yuan, B., (2010) Simplified representation for 3D Point cloud data. *International Conference on Signal Processing Proceedings, ICSP*, pp.1271–1274.
Wang, P., Cao, Y., Shen, C., Liu, L. and Shen, H.T., (2017) Temporal Pyramid Pooling-Based Convolutional Neural Network for Action Recognition. *IEEE Transactions on Circuits and Systems for Video Technology*, 2712, pp.2613–2622.
Wang, X., Xia, M., Cai, H., Gao, Y. and Cattani, C., (2012) Hidden-Markov-Models-based dynamic hand gesture recognition. *Mathematical Problems in Engineering*, 2012.

Wang, X., Zhong, Y., Jin, L. and Xiao, Y., (2022) Scale Adaptive Graph Convolutional Network for Skeleton-Based Action Recognition. *Tianjin Daxue Xuebao (Ziran Kexue yu Gongcheng Jishu Ban)/Journal of Tianjin University Science and Technology*, 553, pp.306–312.
Wu, B., Zhong, J. and Yang, C., (2022) A Visual-Based Gesture Prediction Framework Applied in 107

Social Robots. *IEEE/CAA Journal of Automatica Sinica*, 93, pp.510–519.

Yan, S., Xiong, Y. and Lin, D., (2018) Spatial temporal graph convolutional networks for skeleton-based action recognition. *32nd AAAI Conference on Artificial Intelligence, AAAI 2018*, January, pp.7444–7452.

Yang, H.D., Sclaroff, S. and Lee, S.W., (2009) Sign language spotting with a threshold model based on conditional random fields. *IEEE Transactions on Pattern Analysis and Machine Intelligence*, 317, pp.1264–1277.

Zhang, L., Li, Z., Li, A. and Liu, F., (2018) Large-scale urban point cloud labeling and reconstruction. *ISPRS Journal of Photogrammetry and Remote Sensing*, [online] 138, pp.86–100. Available at: https://doi.org/10.1016/j.isprsjprs.2018.02.008.

Zhang, W., Lin, Z., Cheng, J., Ma, C., Deng, X. and Wang, H., (2020) STA-GCN: two-stream graph convolutional network with spatial–temporal attention for hand gesture recognition. *Visual Computer*, 3610–12, pp.2433–2444.

Zhao, C., Zhou, W., Lu, L. and Zhao, Q., (2019) POOLING SCORES OF NEIGHBORING POINTS FOR IMPROVED 3D POINT CLOUD SEGMENTATION Chenxi Zhao , Weihao Zhou , Li Lu , Qijun Zhao ∗ National Key Laboratory of Fundamental Science on Synthetic Vision , College of Computer Science , Sichuan University , Chengdu . *Proceedings - International Conference on Image Processing, ICIP*, 2019-Septe, pp.1475–1479.

Zhu, C., Yang, J., Shao, Z. and Liu, C., (2021) Vision Based Hand Gesture Recognition Using 3D Shape Context. *IEEE/CAA Journal of Automatica Sinica*, 89, pp.1600–1613.

(Yang et al., 2009; Wang et al., 2012; Celebi et al., 2013; Lin et al., 2014, 2019; Du et al., 2015; Nishida and Nakayama, 2016; Shih and Ma, 2018; Hoang et al., 2019; Li et al., 2019; Nguyen et al., 2019b; a; Ikram and Liu, 2020; Min et al., 2020)

**Bibliography**

1. Neha Baranwal, Ganesh Jaiswal, and Gora Chand Nandi. A speech recognition technique using mfcc with dwt in isolated hindi words. In Intelligent Computing, Networking, and Informatics, pages 697–703. Springer, 2014. Conflict resolution in human-robot interaction.
2. Neha Baranwal and Gora Chand Nandi. An efficient gesture based humanoid learning using wavelet descriptor and mfcc techniques. International Journal of Machine Learning and Cybernetics, 8(4):1369–1388, 2017.
3. Neha Baranwal and Gora Chand Nandi. A mathematical framework for possibility theory-based hidden markov model. International Journal of Bio-Inspired Computation, 10(4):239–247, 2017.
4. Neha Baranwal, Gora Chand Nandi, and Avinash Kumar Singh. Real-time gesture–based communication using possibility theory–based hidden markov model. Computational Intelligence, 33(4):843–862, 2017.
5. Neha Baranwal, Avinash Kumar Singh, and Suna Bench. Extracting primary objects and spatial relations from sentences. In 11th International Conference on Agents and Artificial Intelligence, Prague, Czech Republic, 2019.
6. Neha Baranwal, Avinash Kumar Singh, and Thomas Hellstrom. Fusion of gesture and speech for increased accuracy in human robot interaction. In 2019 24th International Conference on Methods and Models in Automation and Robotics (MMAR), pages 139–144. IEEE, 2019.
7. Neha Baranwal, Avinash Kumar Singh, and Gora Chand Nandi. Development of a framework for human–robot interactions with indian sign language using possibility theory. International Journal of Social Robotics, 9(4):563–574, 2017.
8. Neha Baranwal, Neha Singh, and Gora Chand Nandi. Implementation of mfcc based hand gesture recognition on hoap-2 using webots platform. In 2014 International Conference on Advances in Computing, Communications and Informatics (ICACCI), pages 1897–1902. IEEE, 2014.
9. Neha Baranwal, Kumud Tripathi, and GC Nandi. Possibility theory based continuous indian sign language gesture recognition. In TENCON 2015-2015 IEEE Region 10 Conference, pages 1–5. IEEE, 2015.

10. Neha Baranwal, Shweta Tripathi, and Gora Chand Nandi. A speaker invariant speech recognition technique using hfcc features in isolated hindi words. International Journal of Computational Intelligence Studies, 3(4):277–291, 2014.
11. Avinash Kumar Singh, Neha Baranwal, Kai-Florian Richter, Thomas Hellstrom, and Suna Bensch. Towards verbal explanations by collaborating robot teams. In International Conference on Social Robotics (ICSR19), Workshop Quality of Interaction in Socially Assistive Robots, Madrid, Spain, November 26-29, 2019., 2019.
12. Avinash Kumar Singh, Neha Baranwal, and Gora Chand Nandi. Human perception based criminal identification through human robot interaction. In 2015 Eighth International Conference on Contemporary Computing (IC3), pages 196–201. IEEE, 2015.
13. Avinash Kumar Singh, Neha Baranwal, and Gora Chand Nandi. Development of a self reliant humanoid robot for sketch drawing. Multimedia Tools and Applications, 76(18):18847–18870, 2017.
14. Avinash Kumar Singh, Neha Baranwal, and Gora Chand Nandi. A rough set based reasoning approach for criminal identification. International Journal of Machine Learning and Cybernetics, 10(3):413–431, 2019.
15. Avinash Kumar Singh, Neha Baranwal, and Kai-Florian Richter. An empirical review of calibration techniques for the pepper humanoid robots rgb and depth camera. In Proceedings of SAI Intelligent Systems Conference, pages 1026–1038. Springer, 2019.
16. Avinash Kumar Singh, Pavan Chakraborty, and GC Nandi. Sketch drawing by nao humanoid robot. In TENCON 2015-2015 IEEE Region 10 Conference, pages 1–6. IEEE, 2015.
17. Avinash Kumar Singh, Piyush Joshi, and Gora Chand Nandi. Face liveness detection through face structure analysis. International Journal of Applied Pattern Recognition, 1(4):338–360, 2014.
18. Avinash Kumar Singh, Piyush Joshi, and Gora Chand Nandi. Face recognition with liveness detection using eye and mouth movement. In 2014 International Conference on Signal Propagation and Computer Technology (ICSPCT 2014), pages 592–597. IEEE, 2014.
19. Avinash Kumar Singh, Piyush Joshi, and Gora Chand Nandi. Development of a fuzzy expert system based liveliness detection scheme for biometric authentication. arXiv preprint arXiv:1609.05296, 2016.
20. Avinash Kumar Singh, Arun Kumar, GC Nandi, and Pavan Chakroborty. Expression invariant fragmented face recognition. In 2014 International Conference on Signal Propagation and Computer Technology (ICSPCT 2014), pages 184–189. IEEE, 2014.
21. Avinash Kumar Singh and Gora Chand Nandi. Face recognition using facial symmetry. In Proceedings of the Second International Conference on Computational Science, Engineering and Information Technology, pages 550–554. ACM, 2012.
22. Avinash Kumar Singh and Gora Chand Nandi. Nao humanoid robot: Analysis of calibration techniques for robot sketch drawing. Robotics and Autonomous Systems, 79:108–121, 2016.
23. Avinash Kumar Singh and Gora Chand Nandi. Visual perception-based criminal identification: a query-based approach. Journal of Experimental & Theoretical Artificial Intelligence, 29(1):175–196, 2017.
24. Neha Singh, Neha Baranwal, and GC Nandi. Implementation and evaluation of dwt and mfcc based isl gesture recognition. In 2014 9th International Conference on Industrial and Information Systems (ICIIS), pages 1–7. IEEE, 2014.
25. Kumud Tripathi, Neha Baranwal, and Gora Chand Nandi. Continuous dynamic Indian sign language gesture recognition with invariant backgrounds. In 2015 International Conference on Advances in Computing, Communications and Informatics (ICACCI), pages 2211–2216. IEEE, 2015.
26. Shweta Tripathy, Neha Baranwal, and GC Nandi. A mfcc based hindi speech recognition technique using htk toolkit. In 2013 IEEE Second International Conference on Image Information Processing (ICIIP-2013), pages 539–544. IEEE, 2013.
27. Singh, Avinash Kumar, Baranwal, Neha, Richter, Kai-Florian, Hellström, Thomas and Bensch, Suna. "Verbal explanations by collaborating robot teams" Paladyn, Journal of Behavioral Robotics, vol. 12, no. 1, 2021, pp. 47-57.
28. Singh A.K., Baranwal N., Richter KF., Hellström T., Bensch S. (2020) Understandable Collaborating Robot Teams. In: De La Prieta F. et al. (eds) Highlights in Practical Applications of Agents, Multi-Agent Systems, and Trust-worthiness. The PAAMS Collection. PAAMS 2020. Communications in Computer and Information Science, vol 1233. Springer

29. Singh A.K., Baranwal N., Richter KF., Hellström T., Bensch S. (2020) Understandable Teams of Pepper Robots. In: Demazeau Y., Holvoet T., Corchado J., Costantini S. (eds) Advances in Practical Applications of Agents, Multi-Agent Systems, and Trustworthiness. The PAAMS Collection. PAAMS 2020. Lecture Notes in Computer Science, vol 12092. Springer.
30. Singh, Avinash, Neha Baranwal, and Kai-Florian Richter. "A Fuzzy Inference System for a Visually Grounded Robot State of Mind." 24th European Conference on Artificial Intelligence (ECAI 2020), Including 10th Conference on Prestigious Applications of Artificial Intelligence (PAIS 2020), Virtual, August 29-September 8, 2020. IOS Press, 2020.
31. Baranwal, Neha, and Kamalika Dutta. "Peak detection based spread spectrum Audio Watermarking using discrete Wavelet Transform." International Journal of Computer Applications 24.1 (2011): 16-20.
32. Gadanayak, Bismita, Chittaranjan Pradhan, and Neha Baranwal. "Secured partial MP3 encryption technique." International Journal of Computer Science and Information Technologies 2.4 (2011): 1584-1587.
33. Baranwal, Neha, and Kamalika Datta. "Comparative study of spread spectrum based audio watermarking techniques." 2011 International Conference on Recent Trends in Information Technology (ICRTIT). IEEE, 2011.